\documentclass[a4paper,10pt,twocolumn,conference,dvipsnames]{IEEEtran}

\usepackage[export]{adjustbox}
\usepackage{algorithm}
\usepackage[%
    indLines=false,
    italicComments=true,
    noEnd=false,
]{algpseudocodex}
\usepackage{amsmath}
\usepackage[font=footnotesize,labelfont=bf]{caption}
\usepackage{circledsteps}
\usepackage{csquotes}
\usepackage{datetime2}
\usepackage{fancyhdr}
\usepackage{footmisc}
\usepackage[%
    toc, 
    record, 
    abbreviations, 
    nonumberlist, 
]{glossaries-extra}
\usepackage{graphicx}
\usepackage{hyperref}
\usepackage{import}
\usepackage{makecell}
\usepackage[final]{microtype}
\usepackage{multirow}
\usepackage{newtxmath}  
\usepackage{numprint}
\usepackage{pifont}
\usepackage{ragged2e}
\usepackage{siunitx}
\usepackage{stfloats}
\usepackage{svg}
\usepackage{tabularx}
\usepackage{textcomp}
\usepackage{tikz}
\usepackage{xcolor}

\newcommand{\workname}{ISimDL}

\newcommand{\presentingauthoremail}{alessio.colucci@tuwien.ac.at}

\newcommand{\presentingauthorwebpage}{https://github.com/Alexei95}

\usepackage[%
    backend=biber,%
    backref=false,%
    citestyle=numeric,%
    dashed=false,%
    date=iso,%
    doi=false,%
    giveninits=true,%
    hyperref=true,%
    isbn=false,%
    maxbibnames=100,%
    maxcitenames=100,%
    minbibnames=100,%
    mincitenames=1,%
    seconds=true,%
    sortcites=true,%
    style=ieee,%
    sorting=none,%
    url=false,%
    urldate=iso,%
]{biblatex}

\usepackage{enumitem}

\def\titlefontsize{\fontsize{24.88}{29.856}\selectfont}

\hypersetup{
    colorlinks=true,
    citecolor=.,
    linkcolor=.,
    urlcolor={ForestGreen},
}

\svgsetup{%
    inkscapearea=drawing,
    inkscapepath=svgsubdir,
    inkscapeformat=pdf, 
    inkscapelatex=false, 
    inkscapeopt=-T,  
}

\svgpath{{images/vector/}}

\graphicspath{{images/raster/}{images/vector/exported-pdf/}}

\npthousandsep{\,}

\robustify{\glsentrytitlecase}
\robustify{\gls}
\robustify{\glsxtrshort}
\robustify{\glsxtrshortpl}
\robustify{\GlsXtrIfUnusedOrUndefined}
\robustify{\glsunset}

\newglossary[]{ignored}{}{}{Ignored}
\newglossary[]{conference}{}{}{Conferences}
\newglossary[]{journal}{}{}{Journals}
\newglossary[]{terminology}{}{}{Terminology}

\newabbreviation[
    type={ignored}
]{eo}{EO}{expected outcome}
\newabbreviation[
    type={ignored}
]{nyuad}{NYU Abu Dhabi}{New York University Abu Dhabi, Abu Dhabi, UAE}
\newabbreviation[
    type={ignored}
]{pe}{PE}{proficiency evaluation}
\newabbreviation[
    type={ignored}
]{rq}{RQ}{research question}
\newabbreviation[
    type={ignored}
]{so}{SO}{scientific objective}
\newabbreviation[
    type={ignored}
]{tuwien}{TUWien}{Technische Universit\"at Wien, Vienna, Austria}

\newabbreviation[
    type={terminology}
]{asic}{ASIC}{application specific integrated circuit}
\newabbreviation[
    type={terminology}
]{cnn}{CNN}{convolutional neural network}
\newabbreviation[
    type={terminology}
]{comprnn}{ComprNN}{compressed neural network}
\newabbreviation[
    type={terminology}
]{cps}{CPS}{cyber physical system}
\newabbreviation[
    type={terminology}
]{cpu}{CPU}{central processing unit}
\newabbreviation[
    type={terminology}
]{dl}{DL}{deep learning}
\newabbreviation[
    type={terminology}
]{dmr}{DMR}{dual modular redundancy}
\newabbreviation[
    type={terminology}
]{dnn}{DNN}{deep neural network}
\newabbreviation[
    type={terminology}
]{dre}{DRE}{detectable recoverable error}
\newabbreviation[
    type={terminology}
]{due}{DUE}{detectable unrecoverable error}
\newabbreviation[
    type={terminology}
]{fat}{FAT}{fault aware training}
\newabbreviation[
    type={terminology}
]{fit}{FIT}{failure in time}
\newabbreviation[
    type={terminology}
]{flops}{FLOPs}{floating point operations per second}
\newabbreviation[
    type={terminology}
]{fsd}{FSD}{full self driving}
\newabbreviation[
    type={terminology}
]{gpu}{GPU}{graphics processing unit}
\newabbreviation[
    type={terminology}
]{iot}{IoT}{internet of things}
\newabbreviation[
    type={terminology}
]{lsb}{LSB}{least significant bit}
\newabbreviation[
    type={terminology}
]{ml}{ML}{machine learning}
\newabbreviation[
    type={terminology}
]{msb}{MSB}{most significant bit}
\newabbreviation[
    type={terminology}
]{nan}{NaN}{not a number}
\newabbreviation[
    type={terminology}
]{nn}{NN}{neural network}
\newabbreviation[
    type={terminology}
]{relu}{ReLU}{rectified linear unit}
\newabbreviation[
    type={terminology}
]{risc}{RISC}{reduced instruction set computer}
\newabbreviation[
    type={terminology}
]{sdc}{SDC}{silent data corruption}
\newabbreviation[
    type={terminology}
]{simd}{SIMD}{single instruction multiple data}
\newabbreviation[
    type={terminology}
]{sota}{SotA}{state of the art}
\newabbreviation[
    type={terminology}
]{snn}{SNN}{spiking neural network}
\newabbreviation[
    type={terminology}
]{svm}{SVM}{support vector machine}
\newabbreviation[
    type={terminology}
]{tmr}{TMR}{triple modular redundancy}
\newabbreviation[
    type={terminology}
]{tpu}{TPU}{tensor processing unit}
\newabbreviation[
    type={terminology}
]{ure}{URE}{undetectable recoverable error}
\newabbreviation[
    type={terminology}
]{uue}{UUE}{undetectable unrecoverable error}
\newabbreviation[
    type={terminology}
]{vae}{VAE}{variational auto encoder}

\newabbreviation[
    type={conference}
]{dac}{DAC}{Design Automation Conference}
\newabbreviation[
    type={conference}
]{date}{DATE}{Design Automation and Test in Europe Conference and Exhibition}
\newabbreviation[
    type={conference}
]{dsn}{DSN}{International Conference on Dependable Systems and Networks}
\newabbreviation[
    type={conference}
]{iccad}{ICCAD}{International Conference on Computer-Aided Design}
\newabbreviation[
    type={conference}
]{iclr}{ICLR}{International Conference on Learning Representations}
\newabbreviation[
    type={conference}
]{icml}{ICML}{International Conference on Machine Learning}
\newabbreviation[
    type={conference}
]{ijcnn}{IJCNN}{IEEE International Joint Conference on Neural Networks}
\newabbreviation[
    type={conference}
]{iros}{IROS}{IEEE/RJS International Conference on Intelligent Robots and Systems}
\newabbreviation[
    type={conference}
]{nips}{NeurIPS}{Neural Information Processing Systems}
\newabbreviation[
    type={journal}
]{dandt}{D\&T}{IEEE Design \& Test}
\newabbreviation[
    type={journal}
]{jmlr}{JMLR}{Journal of Machine Learning Research}
\newabbreviation[
    type={journal}
]{tcad}{TCAD}{IEEE Transaction on Computer-Aided Design of Integrated Circuits and Systems}
\newabbreviation[
    type={journal}
]{tecs}{TECS}{ACM Transactions on Embedded Computing Systems}
\newabbreviation[
    type={journal}
]{tvlsi}{TVLSI}{IEEE Transactions on Very Large Scale Integration (VLSI) Systems}

\glsaddall

\glssetcategoryattribute{abbreviation}{glossdesc}{title}

\setkeys{glslink}{hyper=false}

\newcommand{\emphasis}[1]{\emph{#1}}

\newcommand{\emphasizedworkname}{\emphasis{\workname}}

\newcommand{\esm}[1]{\ensuremath{#1}}

\newcommand{\ms}[1]{\esm{\mathsf{#1}}}

\newcommand{\customcheckmark}{\ding{51}}
\newcommand{\customxmark}{\ding{55}}
\newcommand*\colourmark[3]{%
  \expandafter\newcommand\csname #3\endcsname{\textcolor{#1}{#2}}}

\newcommand{\red}[1]{\textcolor{red}{#1}}

\makeatletter
\newcommand{\printfnsymbol}[1]{%
    \textsuperscript{\@fnsymbol{#1}}%
}
\makeatother

\makeatletter
\newcommand*{\inlineequation}[2][]{%
  \begingroup
    \refstepcounter{equation}%
    \ifx\\#1\\%
    \else
      \label{#1}%
    \fi
    \relpenalty=10000 %
    \binoppenalty=10000 %
    \ensuremath{%
      #2%
    }%
    ~\@eqnnum
  \endgroup
}
\makeatother

\definecolor{blanchedalmond}{rgb}{1.0, 0.92, 0.8}
\definecolor{champagne}{rgb}{0.97, 0.91, 0.81}
 	\definecolor{beige}{rgb}{0.96, 0.96, 0.86}
 	\definecolor{carnelian}{rgb}{0.7, 0.11, 0.11}
\definecolor{crimson}{rgb}{0.86, 0.08, 0.24}
\definecolor{yellowcirclefill}{RGB}{255, 246, 221}
\definecolor{redcircleborder}{RGB}{192, 0, 0}
\definecolor{inkscapered}{RGB}{255, 0, 0}
\definecolor{inkscapedarkgreen}{RGB}{0, 128, 0}

\newcommand{\RedCircled}[1]{\Circled[inner color=white,outer color=inkscapered,fill color=inkscapered]{#1}}
\newcommand{\GoodRedCircled}[1]{\RedCircled{\textbf{\small{#1}}}}
\newcommand{\DarkGreenCircled}[1]{\Circled[inner color=white,outer color=inkscapedarkgreen,fill color=inkscapedarkgreen]{#1}}
\newcommand{\GoodDarkGreenCircled}[1]{\DarkGreenCircled{\textbf{\small{#1}}}}

\newcommand{\titlecaseabbreviation}[1]{\GlsXtrIfUnusedOrUndefined{#1}{\glsentrytitlecase{#1}{long}\space(\glsxtrshort{#1})\glsunset{#1}}{\glsxtrshort{#1}\glsunset{#1}}}
\newcommand{\titlecaseabbreviationpl}[1]{\GlsXtrIfUnusedOrUndefined{#1}{\glsentrytitlecase{#1}{longpl}\space(\glsxtrshortpl{#1})\glsunset{#1}}{\glsxtrshortpl{#1}\glsunset{#1}}}

\let\svthefootnote\thefootnote
\newcommand\blankfootnote[1]{%
  \let\thefootnote\relax\footnotetext{#1}%
  \let\thefootnote\svthefootnote%
}
\let\svfootnote\footnote
\renewcommand\footnote[2][?]{%
  \if\relax#1\relax%
    \blankfootnote{#2}%
  \else%
    \if?#1\svfootnote{#2}\else\svfootnote[#1]{#2}\fi%
  \fi
}

\let\svfootnote\footnote

\let\svfootnote\footnote

\begin{document}

\colourmark{green}{\customcheckmark}{greencheckmark}
\colourmark{red}{\customxmark}{redxmark}
\colourmark{red}{\customcheckmark}{redcheckmark}
\colourmark{green}{\customxmark}{greenxmark}

\title{\titlefontsize \workname{}: Importance Sampling-Driven Acceleration of Fault Injection Simulations for Evaluating the Robustness of \glsentrytitlecase{dl}{long}}

\author{%
\IEEEauthorblockN{Alessio Colucci$^{1}$, Andreas Steininger$^{1}$, Muhammad Shafique$^{2}$}
\IEEEauthorblockA{\textit{$^1$Technische Universit{\"a}t Wien, Vienna, Austria}} 
\IEEEauthorblockA{\textit{$^2$Division of Engineering, New York University Abu Dhabi, UAE}}\vspace*{-4mm}\\
Email: \{alessio.colucci,andreas.steininger\}@tuwien.ac.at,muhammad.shafique@nyu.edu\\
}




\maketitle

\begin{abstract}
\glslocalresetall \titlecaseabbreviation{dl} systems have proliferated in many applications, requiring specialized hardware accelerators and chips. In the nano-era, devices have become increasingly more susceptible to permanent and transient faults. Therefore, we need an efficient methodology for analyzing the resilience of advanced \titlecaseabbreviation{dl} systems against such faults, and understand how the faults in neural accelerator chips manifest as errors at the \titlecaseabbreviation{dl} application level, where faults can lead to undetectable and unrecoverable errors.
Using fault injection, we can perform resilience investigations of the \titlecaseabbreviation{dl} system by modifying neuron weights and outputs at the software-level, as if the hardware had been affected by a transient fault. Typically, the way the faults are chosen for such fault injection is based on random uniform sampling of the space of all the possible faults. However, this method of random uniform sampling of neurons and faults is extremely inefficient for large \titlecaseabbreviationpl{dnn}, leading to huge inaccuracies and slow execution of fault injection experiments. Existing fault models reduce the search space, allowing faster analysis, but requiring a-priori knowledge on the model, and not allowing further analysis of the filtered-out search space.
Therefore, we propose \emphasizedworkname{}, a novel methodology that employs neuron sensitivity to generate importance sampling-based fault-scenarios. Without any a-priori knowledge of the model-under-test, \emphasizedworkname{} provides an equivalent reduction of the search space as existing works, while allowing long simulations to cover all the possible faults, improving on existing model requirements.
Our experiments show that the importance sampling provides up to \num{15}$\times$ higher precision in selecting critical faults than the random uniform sampling, reaching such precision in less than \num{100} faults. Additionally, we showcase another practical use-case for importance sampling for reliable \titlecaseabbreviation{dnn} design, namely \emph{\titlecaseabbreviation{fat}}. By using \emphasizedworkname{} to select the faults leading to errors, we can insert the faults during the \titlecaseabbreviation{dnn} training process to harden the \titlecaseabbreviationpl{dnn} against such faults. Using importance sampling in \titlecaseabbreviation{fat} reduces the overhead required for finding faults that lead to a predetermined drop in accuracy by more than \num{12}$\times$.

\glslocalresetall

\end{abstract}


\glsresetall


\section{Introduction}
\label{section:introduction}

In recent years, \titlecaseabbreviation{dl} has become common in many fields, such as medical imaging, autonomous driving and voice recognition \cite{dongSurveyDeepLearning2021}, requiring specialized hardware accelerators and chips \cite{szeEfficientProcessingDeep2020}.
In the nano-era, error resilience has increased in importance as technology scaling leads to more faults happening per transistor \cite{nealeNeutronRadiationInduced2016}, and newer applications require higher safety standard than before \cite{internationalorganizationforstandardizationISO2626212018}. Companies have tried to approach error resilience \cite{lotfiResiliencyAutomotiveObject2019a,zhangEnablingTimingError2020}, however, it is difficult to strike a balance between efficiency and error resilience, as shown in examples like \cite{FSDChipTesla}, where the implemented solution is the expensive \titlecaseabbreviation{dmr}.
Therefore, many research works have focused on improving error analysis tools and mitigation techniques. One of such analysis tools is fault injection, purposely injecting faults in the execution of \titlecaseabbreviationpl{dnn} to verify whether the injected fault leads to an error, and develop appropriate mitigation techniques. Many \gls{sota} works have developed tools to increase fault injection simulation performance \cite{liTensorFIConfigurableFault2018,mahmoudPyTorchFIRuntimePerturbation2020,colucciEnpheephFaultInjection2022a,agarwalLLTFIFrameworkAgnostic2022}, or to decrease the fault search space \cite{chenBinFIEfficientFault2019,jhaMLBasedFaultInjection2019}, and finally to build on these advancements to generate new mitigation techniques \cite{chenLowcostFaultCorrector2021}.
Decreasing the fault search space will speed up fault injection simulations, while making the error model less accurate and effective, as faults which have been discarded might be leading to errors. This is shown in \gls{sota}, as they achieve only \num{80}\% precision for critical faults, a measure of the efficacy of the fault sampling\footnotemark, while still requiring hundreds of thousands of faults, instead of quadrillions in the whole search space \cite{chenBinFIEfficientFault2019,jhaMLBasedFaultInjection2019}.
Additionally, current state-of-the-art fault-sampling models reduce the fault search space by employing human-made fault models, hence requiring the external a-priori knowledge to select the fault search space.
\emphasizedworkname{} aims to increase the effectiveness and efficiency of the fault injection campaign while still covering the whole fault search space over long-running simulations.

\footnotetext{Precision and recall, two important metrics for measuring efficacy and effectiveness of fault injection, will be mathematically defined and explained in Section \ref{section:experimental_setup:subsection:metrics}}

\subsection{Motivational Case Study}
\label{section:introduction:subsection:motivationalcasestudy}

\begin{figure}[htbp]
    \centering
    \includegraphics[width=0.7\linewidth]{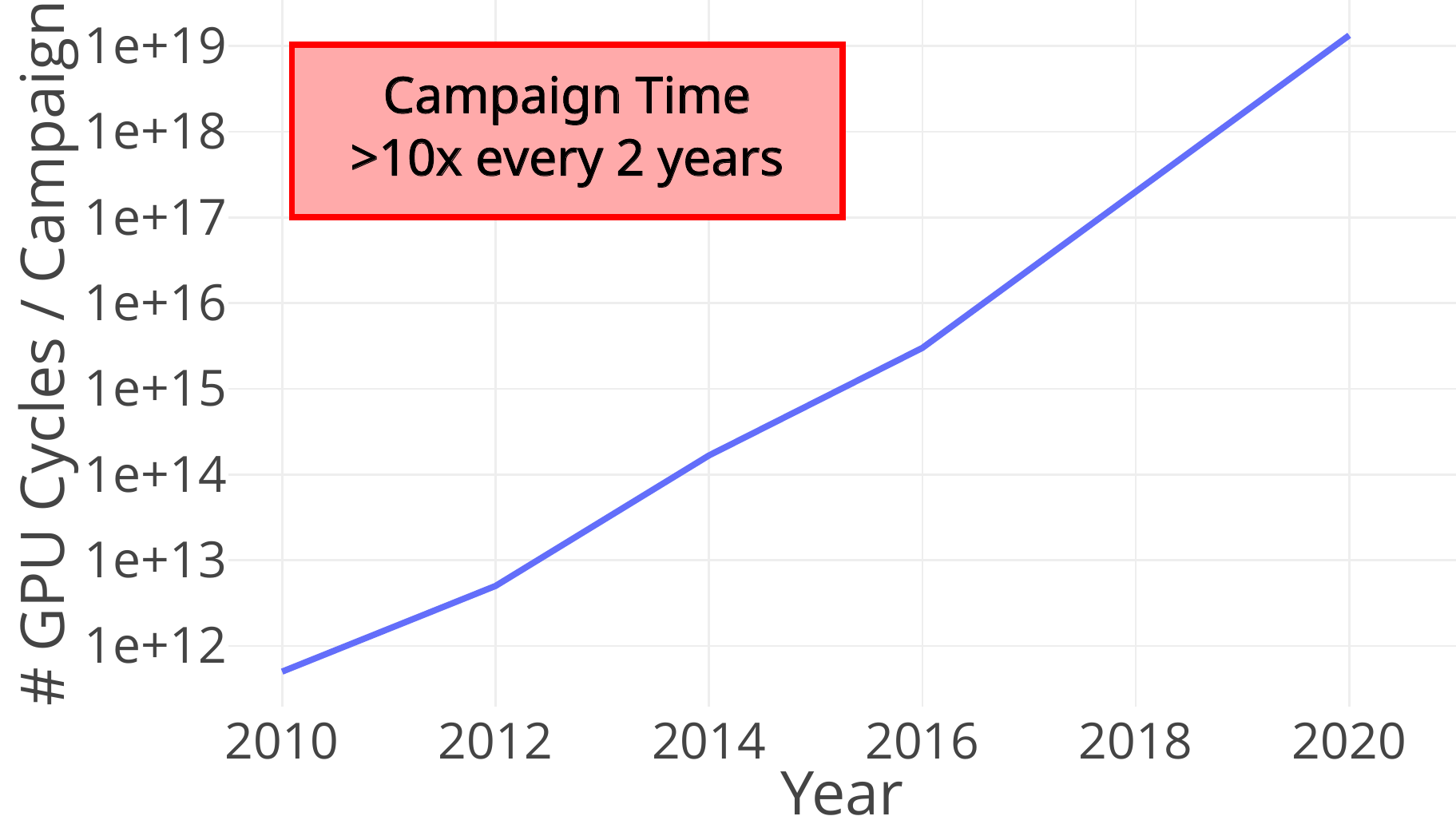}
    \caption{Trend of number of \titlecaseabbreviation{gpu} cycles required to perform a complete fault injection campaign. The trend is exponential, increasing by more than one order of magnitude every 2 years.}
    \label{figure:motivational_case_study:subfigure:quantitative_campaign_duration}
\end{figure}
\begin{figure}[htbp]
    \centering
    \includegraphics[width=0.7\linewidth]{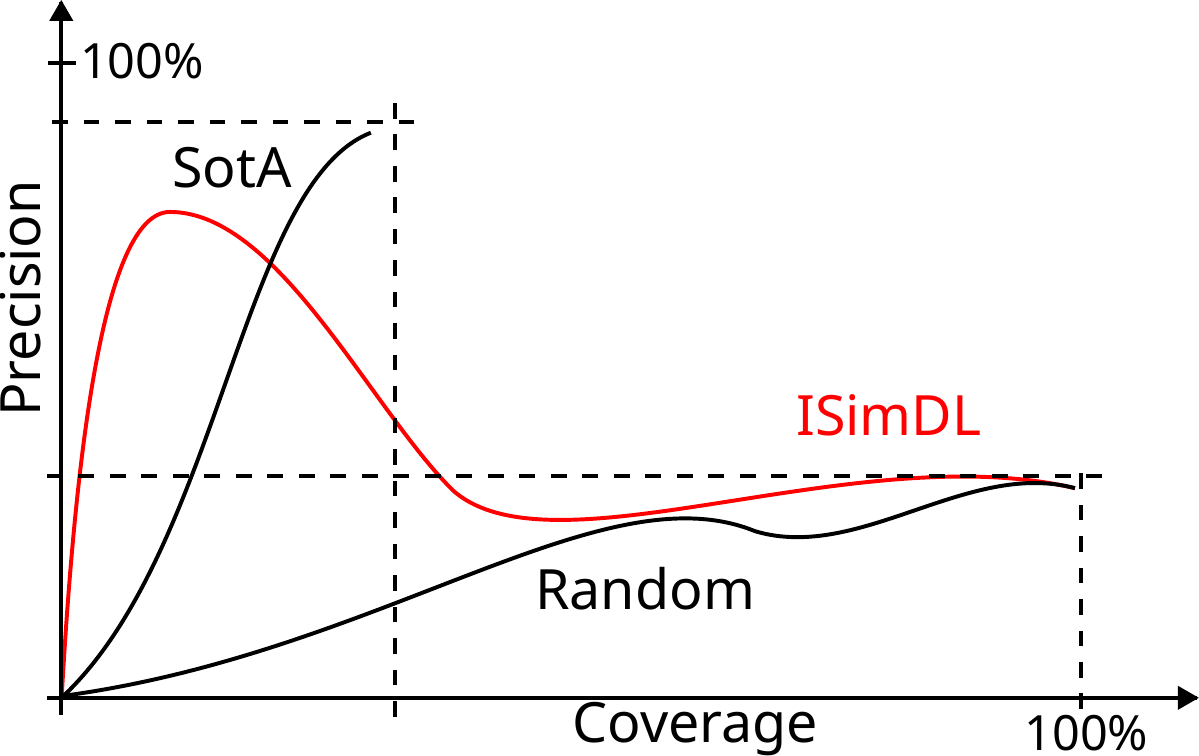}
    \caption{Qualitative comparison among \gls{sota}, random sampling and \emphasizedworkname{}. \gls{sota} reaches higher precision but it does not cover all the possible faults, random sampling has very low precision but can cover all the faults, while \emphasizedworkname{} has very high precision with the first few samples while reaching random sampling level of coverage, merging the best of both worlds.}
    \label{figure:motivational_case_study:subfigure:qualitative_coverage_precision_plot}
\end{figure}

In Figure \ref{figure:motivational_case_study:subfigure:quantitative_campaign_duration}, we merge the latest trends in terms of \titlecaseabbreviation{dnn} parameter count \cite{villalobosMachineLearningModel2022} and required execution \titlecaseabbreviation{flops} \cite{desislavovComputeEnergyConsumption2021}, hardware transistor count \cite{sunSummarizingCPUGPU2020,owidtechnologicalchange} and generated hardware \titlecaseabbreviation{flops} \cite{sunSummarizingCPUGPU2020}, to show the approximate number of \titlecaseabbreviation{gpu} cycles required to perform a full injection campaign on \gls{sota} \titlecaseabbreviation{dnn} models. We can see the exponential trend, increasing by more than one order of magnitude every 2 years. \emph{This highlights the necessity of efficiently selecting the faults to be injected, as it is infeasible to analyze all the possible faults.} 

In Figure \ref{figure:motivational_case_study:subfigure:qualitative_coverage_precision_plot}, we show a comparison between \gls{sota} against random fault sampling. We can see how \titlecaseabbreviation{sota} reaches high precision quickly, around \num{80}\%, however, it covers a very small percentage of the whole fault search space, roughly \num{20}\%. On the other hand, random fault sampling covers all the reachable fault search space, but it has very low precision as most of the faults do not lead to errors. Therefore, we need a new fault-sampling model to achieve very high precision with the first few samples, but still capable of covering the remaining fault search space if required, without using any external a-priori knowledge. \emph{\workname{} manages to achieve this balance, reaching up to \num{60}\% precision with less than \num{100} faults.}

\subsection{Research Questions}
\label{section:introduction:subsection:research_questions}

The aforementioned case study leads to the following research questions:

\begin{itemize}
    \item How can we provide a fault model which is as precise as state-of-the-art for the initial samples, while covering the whole search space, and without requiring a-priori knowledge?
    \item How do the different importance sampling algorithms compare? Which one is best suited to achieve \gls{sota} performance?
    \item How can importance sampling algorithms benefit specific resilience applications, e.g., \titlecaseabbreviation{fat}?
\end{itemize}

\subsection{Novel Contributions}
\label{section:introduction:subsection:novel_contributions}

\begin{figure}[h]
    \centering
    \includegraphics[width=\linewidth]{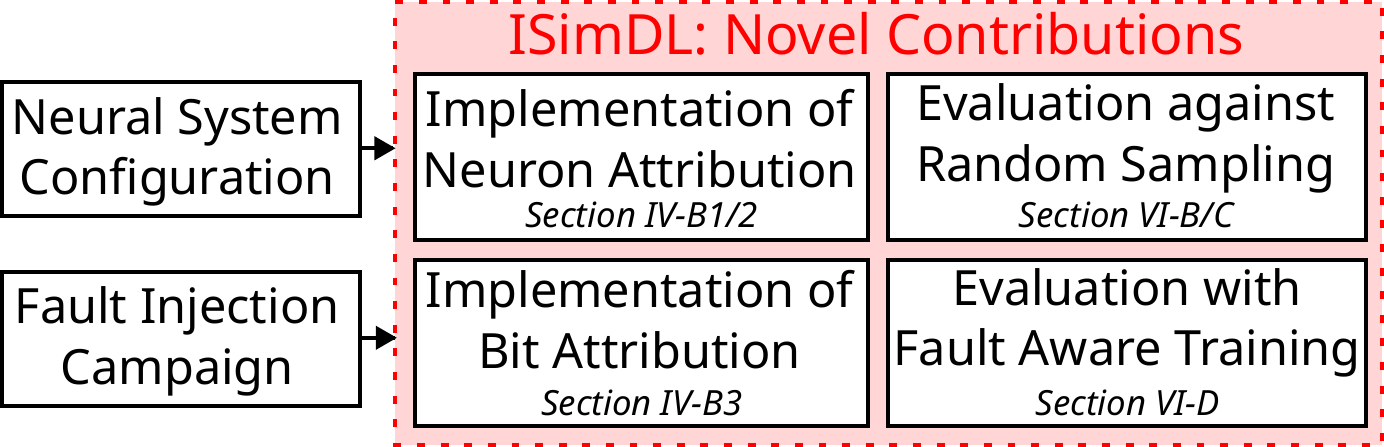}
    \caption{Novel contributions presented in \emphasizedworkname{}.}
    \label{figure:novel_contributions}
\end{figure}

To answer the research questions, we provide the following novel contributions in \emphasizedworkname{}, as shown in Figure \ref{figure:novel_contributions}:

\begin{itemize}
    \item we develop and implement an importance sampling algorithm for computing the importance of both neuron weights and neuron outputs (Section \ref{section:methodology:subsection:attribution_computation:subsubsection:neuron_output} \& \ref{section:methodology:subsection:attribution_computation:subsubsection:neuron_weight});
    \item we develop and implement importance sampling attribution for tensor bits (Section \ref{section:methodology:subsection:attribution_computation:subsubsection:tensor_bit});
    \item we test \emphasizedworkname{} on \gls{sota} \titlecaseabbreviationpl{dnn}, VGG11 and ResNet18, trained on different datasets, CIFAR10 and GTSRB, comparing it with random sampling (Section \ref{section:evaluation:subsection:attribution_efficacy} \& \ref{section:evaluation:subsection:different_configurations});
    \item we employ \emphasizedworkname{} to speed up a possible use-case of fault injection, \titlecaseabbreviation{fat} (Section \ref{section:evaluation:subsection:fault_aware_training}).
\end{itemize}

\subsection{Paper Organization}

After presenting background information in Section \ref{section:background} and \gls{sota} in Section \ref{section:related_work}, we show the methodology of \emphasizedworkname{} in Section \ref{section:methodology}, as well as the setup used for the experiments in Section \ref{section:experimental_setup}. We finally present the results and the comparisons with \gls{sota} and random sampling in Section \ref{section:evaluation}, before drawing the conclusions in Section \ref{section:conclusion}.

\section{Background}
\label{section:background}

\subsection{Fault Definitions \& Fault Propagation}
\label{section:background:subsection:fault_defitions_propagation}

\begin{figure}[h]
    \centering
    \includegraphics[width=\linewidth]{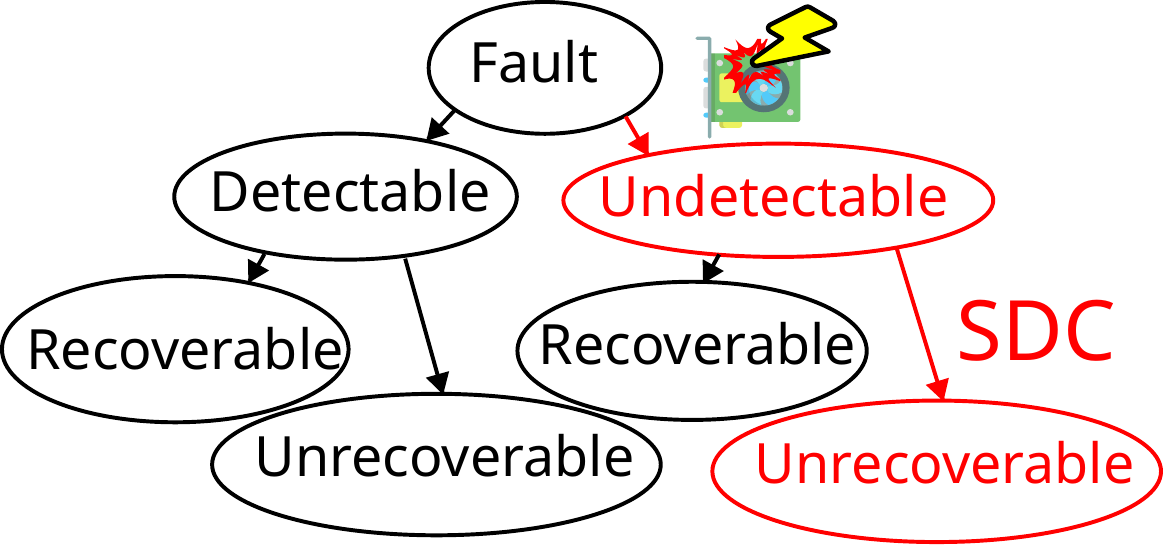}
    \caption{Fault classification based on detectability and recoverability from the resulting modifications: in this work we focus on \glsentrytitlecase{sdc}{longpl}, as they are undetectable unrecoverable errors, leading to huge drops in \glsentryshort{dnn} accuracy.}
    \label{figure:fault_background}
\end{figure}

Faults propagate in different ways depending on the software and hardware conditions in which they occur: if the outcome leads to a change of the processed data only, we refer to them as soft errors. Depending on the outcome on the system, as normal execution or soft errors, faults can be distinguished in 4 categories, as described in Figure \ref{figure:fault_background}: faults can be detectable or undetectable, and the resulting outcome can be recoverable or unrecoverable. Therefore, we can define 4 types of errors: \titlecaseabbreviationpl{dre}, \titlecaseabbreviationpl{due}, \titlecaseabbreviationpl{sdc}, which corresponds to \titlecaseabbreviationpl{uue}, \titlecaseabbreviationpl{ure}. \titlecaseabbreviationpl{dre} and \titlecaseabbreviationpl{ure} are recoverable, hence, they are not analyzed in detail in this work. Regarding \titlecaseabbreviationpl{due}, these errors can be detected through many efficient algorithms, and while they might require complete system reboots, they are detectable, and a possible mitigation can be employed. Therefore, the main focus of this work is for \titlecaseabbreviationpl{sdc} regarding soft errors, as these errors can lead to great drops in accuracy for \titlecaseabbreviation{dnn} systems, and they are not easily predictable. \titlecaseabbreviationpl{sdc} can also affect directly the system execution, and not only its data, however we do not consider them here, hence, we focus only on soft errors, affecting data and the result of the \titlecaseabbreviation{dnn} system. This is mainly because most \titlecaseabbreviationpl{cps} have strict safety requirements, which might be invalidated by possible errors.

Another classification system for faults is transient and permanent: the former refers to faults generated by temporary variations in the running hardware and software system, such as charged particles interacting with the silicon substrate, while the latter refers to permanent defects of the system. We focus on transient faults, as scaling technologies has increased the effects of faults on the running systems. In the case of transient faults caused by charged particles, the resulting effect is a bit flip in the corresponding registers and memory cells. Therefore, we use these transient faults causing bit-flips as our study target.

\subsection{Fault Injection for \glsentrytitlecase{dnn}{longpl}}
\label{section:background:subsection:fault_injection_for_neural_networks}

Fault injection is a tool for testing the response of a system when an unreachable or undesired state is erroneously reached.
For \titlecaseabbreviationpl{dnn}, fault injection can be implemented in many different ways: a common way of injecting faults is by modifying weight and/or output tensors during execution. This approach is the most reliable and the easiest to implement for simulating bit-flips due to charged particle interactions, for more complex faults it might be required to cover control logic crashes or modifications, which are out of the scope of this work. Therefore, we focus only on modifying the weight and the output tensors.

\subsection{Importance Sampling}
\label{section:background:subsection:importance_sampling}

\begin{figure}[h]
    \centering
    \includegraphics[width=0.8\linewidth]{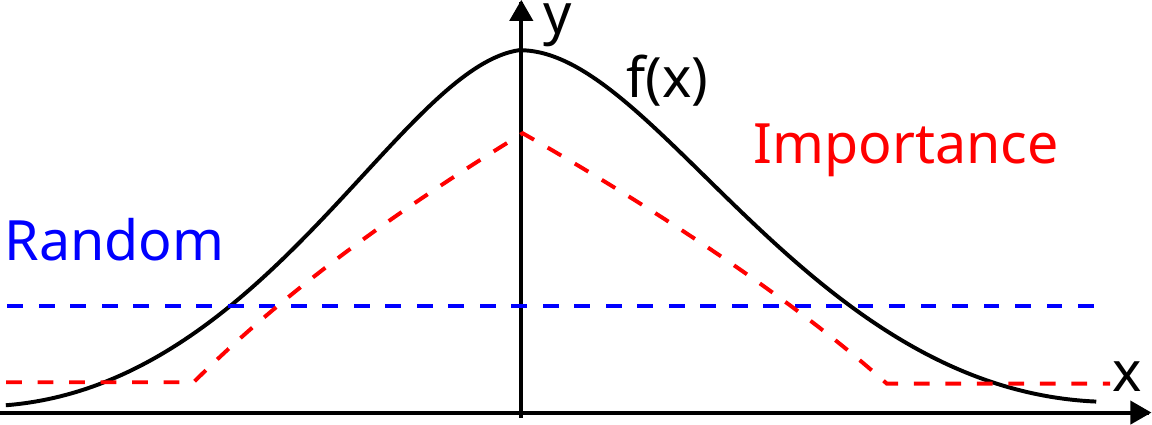}
    \caption{Qualitative visualization of differences between random uniform sampling and importance sampling for integral computation. Importance sampling uses a sampling distribution focused where the integrand function is bigger in absolute value, hence providing more accurate result with fewer samples than random uniform sampling.}
    \label{figure:importance_sampling_background}
\end{figure}

Importance Sampling is a Monte Carlo sampling technique used in integral computations \cite{kloekBayesianEstimatesEquation1978}.
We show an example for integral computation in Figure \ref{figure:importance_sampling_background}. For computing $\int_{-\infty}^{\infty} f\left(x\right)~dx$, we can sample either a uniform distribution which is completely random, or use importance sampling, employing a modified distribution. The latter increases the sampling rate around the areas where $f\left(x\right)$ is larger in absolute value, hence providing a more accurate computation with fewer samples compared to a random uniform distribution.
The application to fault injection is similar, focusing the fault samples where the effects would be more noticeable, compared to random sampling which covers all the possible faults in the same way.

\section{Related Work}
\label{section:related_work}

\begin{table}[h]
    \caption{Comparison between \emphasizedworkname{} and other \gls{sota} works. \emphasizedworkname{} provides the advantage of covering the whole fault search space, while achieving high precision with no a-priori knowledge.}
    \label{table:sota_comparison}
    \centering
    \setlength{\tabcolsep}{2pt}
    \resizebox{\linewidth}{!}{%
        \begin{tabular}{c|c|c|c|c|c|c|}
\cline{2-7}
 &
  \red{\textbf{\emphasizedworkname{}}} &
  \textbf{\begin{tabular}[c]{@{}c@{}}enpheeph\\ \cite{colucciEnpheephFaultInjection2022a}\end{tabular}} &
  \textbf{\begin{tabular}[c]{@{}c@{}}TensorFI\\ \cite{liTensorFIConfigurableFault2018}\end{tabular}} &
  \textbf{\begin{tabular}[c]{@{}c@{}}LLTFI\\ \cite{agarwalLLTFIFrameworkAgnostic2022}\end{tabular}} &
  \textbf{\begin{tabular}[c]{@{}c@{}}BinFI\\ \cite{chenBinFIEfficientFault2019}\end{tabular}} &
  \textbf{\begin{tabular}[c]{@{}c@{}}AVFI\\ \cite{jhaMLBasedFaultInjection2019}\end{tabular}} \\ \hline
\multicolumn{1}{|c|}{\textbf{Sampling Type}} &
  Importance &
  Random &
  Random &
  Random &
  Random &
  Random \\ \hline
\multicolumn{1}{|c|}{\textbf{Search Space}} &
  Full &
  Full &
  Full &
  Full &
  Reduced &
  Reduced \\ \hline
\end{tabular}
    }
\end{table}

There have been many work focusing on fault injection for \titlecaseabbreviationpl{dnn}, focusing on the tools required for fault injection as well as the different sampling models. We show a comparison in Table \ref{table:sota_comparison}.
enpheeph \cite{colucciEnpheephFaultInjection2022a}, TensorFI \cite{liTensorFIConfigurableFault2018} and LLTFI \cite{agarwalLLTFIFrameworkAgnostic2022} focus on providing innovative fault injection frameworks to speed up the overhead of running fault injection compared to normal \titlecaseabbreviation{dnn} execution. However, they do not employ any specific algorithm for sampling, resorting to random uniform sampling. At the same time, they are able to cover the whole search space, as they do not filter the possible faults.
On the other hand, BinFI \cite{chenBinFIEfficientFault2019} and AVFI \cite{jhaMLBasedFaultInjection2019} provide improved fault models, not focusing on how the fault injection is executed as the previous works, but how to choose the faults using external knowledge, in their case human knowledge. However, they still employ random uniform sampling, but as they decrease the extension of the fault search space, they reach higher precision than the previous tools.
\emphasizedworkname{} employs a novel sampling method based on importance sampling, hence it is capable of achieving similar precision while still requiring no human knowledge and without reducing the fault search space.

\section{Methodology}
\label{section:methodology}

\begin{figure}[h]
    \centering
    \includegraphics[width=\linewidth]{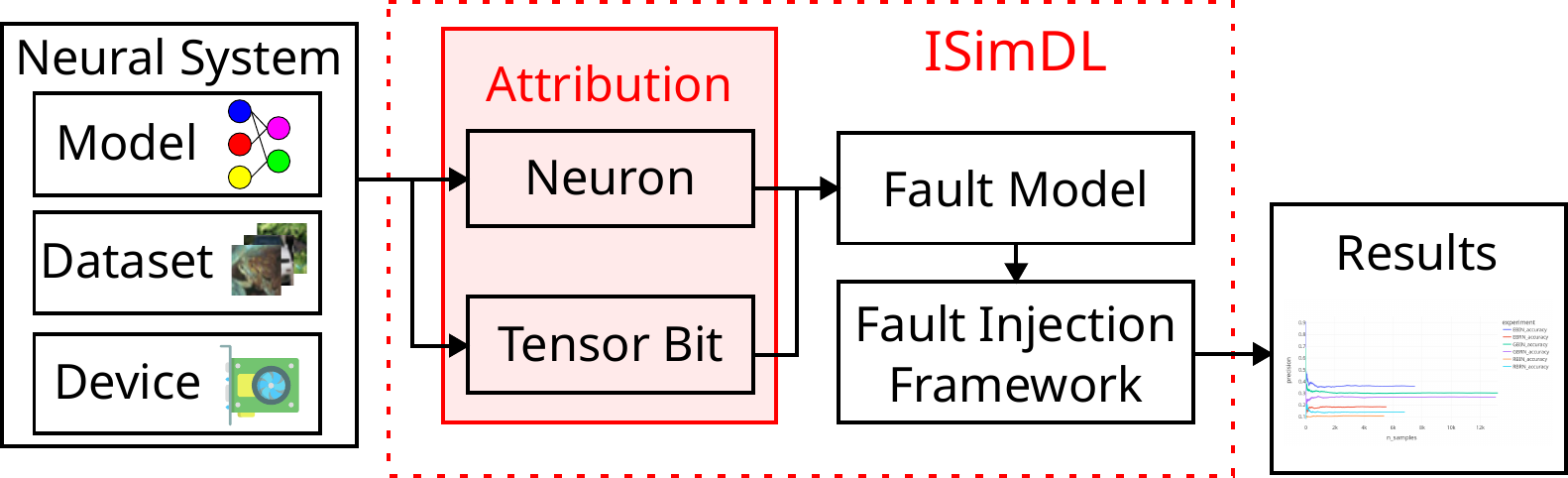}
    \caption{Overview of \emphasizedworkname{}. The required inputs are on the left, while the main contribution is highlighted in red as Attribution. The results are symbolized on the right in the form of plots and statistics on the fault injection campaign.}
    \label{figure:methodology}
\end{figure}

We approach the research questions mentioned in Section \ref{section:introduction:subsection:research_questions}, by developing a novel sampling technique, based on importance sampling. We show an overview of \emphasizedworkname{} in Figure \ref{figure:methodology}. As described in Section \ref{section:background:subsection:fault_injection_for_neural_networks}, we employ fault injections on the tensors corresponding to neuron weight and neuron outputs, as they are the most representative of possible \titlecaseabbreviationpl{sdc}. We describe the rest of the \emphasizedworkname{} methodology in the following sub-sections.

\subsection{Inputs}

\emphasizedworkname{} requires a complete neural system as input, comprising a \titlecaseabbreviation{dnn} model, a dataset and a device optimized for running them.
There are no other inputs required, as the fault selection is done internally by \emphasizedworkname{}, hence not requiring any other human inputs as for other \gls{sota} works.

\subsection{Attribution Computation}

\begin{figure}[h]
    \centering
    \includegraphics[width=0.6\linewidth]{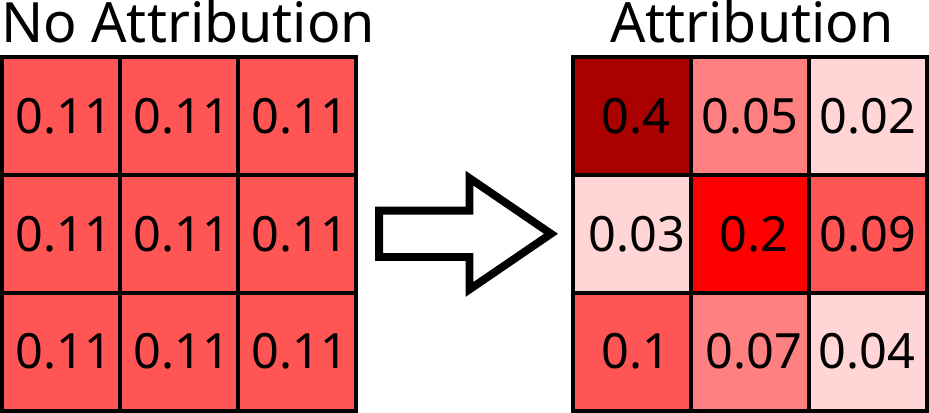}
    \caption{A comparison of sampling faults without and with attribution: when not using attribution, all the target elements have the same weight, hence the random uniform sampling behavior. When using attribution, the importance of each element on the output computation is used as a weight for the sampling algorithm, directing it towards faults with more potential impact.}
    \label{figure:attribution}
\end{figure}

Attribution computation is the main novel contribution provided by \emphasizedworkname{}.
It is used to generate the weights to be used when sampling the faults, as shown in Figure \ref{figure:attribution}. When attribution is not used, each element has the same weight, hence resulting in random uniform sampling. However, attribution takes into account the importance of each element when computing the model output, hence resulting in asymmetric weights for the sampling.
The attribution implementation is different if we are computing the attribution for the neuron weight or the neuron output, as different formulas are used. Additionally, the attribution computation reaches up to each bit in the target tensor.
The implementations are detailed in the following paragraphs.


\subsubsection{Neuron Output}
\label{section:methodology:subsection:attribution_computation:subsubsection:neuron_output}

For the neuron output attribution, we use the layer conductance definition, as defined in \cite{shrikumarComputationallyEfficientMeasures2018,dhamdhereHowImportantNeuron2018}. We report the used equation here:

\begin{equation*}
    \label{equation:weight_conductance}
    \ms{Cond}^{y}(x) ::= \sum_{i}  (x_i-x'_i)\cdot\int_{\alpha=0}^{1} \tfrac{\partial F(x' + \alpha (x-x'))}{\partial y} \cdot \tfrac{\partial y}{\partial x_i} ~d\alpha 
\end{equation*}

$y$ is the output of the neuron, $x$ the test inputs to the network, varying over $i$, $x'$ the baseline input, $F$ is the model function. 
We sum the deviation from the baseline of each input weighted by the integral of the gradient computed between the actual input and the baseline, further weighing the gradient by the gradient of the neuron output with respect to the actual input. In this way we can compute the deviations with respect to a baseline execution, making the attribution closer to the real importance of each neuron output.

\subsubsection{Neuron Weight}
\label{section:methodology:subsection:attribution_computation:subsubsection:neuron_weight}

The attribution for the weight of a neuron is computed with respect to the generated outputs, using the following formula:

\begin{equation*}
    \label{equation:layer_conductance}
    \ms{WeightAttr}_j^L(x) =\sum_{i = 1}^{N} \tfrac{\partial L(x_i)}{\partial w_j}
\end{equation*}

$L$ is the function of the layer we compute the attribution for, $w_j$ is the weight we are computing the attribution for, with $j$ being the index, $x_i$ are the possible inputs used in the gradient computation.
The formula sums all the gradients computed from all the outputs for each weight tensor, and this is used as a weight for the sampling algorithm.
The number of inputs $N$ to be used depends on the application and the specific implementation, but can vary from 1 to a full dataset.

\subsubsection{Tensor Bit}
\label{section:methodology:subsection:attribution_computation:subsubsection:tensor_bit}

\begin{figure}[h]
    \begin{algorithm}[H]
        \begin{footnotesize}
        \captionsetup{font=footnotesize}
        \caption{Gradient Computation for FloatingPoint32 Tensor Bits}
        \label{algorithm:methodology_bit_gradient}
        \begin{algorithmic}[1]
            \Procedure{ComputeBitGradients}{$tensor$, $attribution$}
                \LComment{each bit in the list is a tensor with gradient accumulation}
                \State $list_{bits}$ $\gets$ bits from $tensor$, Least Significant Bit at the 0-index
                \State $bits_{sign}$ $\gets$ Most Significant Bit from $list_{bits}$
                \State $bits_{exponent}$ $\gets$ 8 bits from MSB to LSB from $list_{bits}$
                \State $bits_{mantissa}$ $\gets$ 23 remaining bits, reaching LSB, from $list_{bits}$
                \State $sign$ $\gets$ $-2 \times bits_{sign} + 1$
                \State $exponent$ $\gets$ $-127$
                \For{$index$, $bit$ in $bits_{exponent}$}
                    \State $exponent$ $\gets$ $exponent + 2^{index} \times bit$
                \EndFor
                \State $mantissa$ $\gets$ $1$
                \For{$index$, $bit$ in $bits_{mantissa}$}
                    \State $mantissa$ $\gets$ $mantissa + 2^{index - 23} \times bit$
                \EndFor
                \If{$\ms{all}\left(bits_{exponent} == 0\right)$} \label{algorithm:line:flag_start}
                    \State $flag_{infinity-nan}$ $\gets$ $0$
                    \If{$\ms{all}\left(bits_{mantissa} == 0\right)$}
                        \LComment{zero representations}
                        \State $flag_{notzero}$ $\gets$ $0$
                    \Else
                        \LComment{denormalized numbers}
                        \State $flag_{notzero}$ $\gets$ $1$
                        \State $exponent$ $\gets$ $exponent + 1$
                        \State $mantissa$ $\gets$ $mantissa - 1$
                    \EndIf
                \ElsIf{$\ms{all}\left(bits_{exponent} == 1\right)$}
                    \LComment{infinity or not-a-number representations}
                    \LComment{we cannot use infinite numbers, }
                    \LComment{otherwise gradient would be not-a-number}
                    \State $flag_{notzero}$ $\gets$ $1$
                    \State $flag_{infinity-nan}$ $\gets$ $1$
                \Else
                    \State $flag_{notzero}$ $\gets$ $1$
                    \State $flag_{infinity-nan}$ $\gets$ $0$
                \EndIf \label{algorithm:line:flag_end}
                \State $value$ $\gets$ $sign \times 2^{exponent} \times mantissa$ 
                \State $value$ $\gets$ $value \times flag_{notzero}$ \label{algorithm:line:flag_multiply_zero}
                \If{$flag_{infinity-nan} == 1$}
                    \State $value$ $\gets$ $value \div value \div 33 \times float_{max~representable}$ \label{algorithm:line:flag_multiply_inf}
                \EndIf
                \State $value.grad$ $\gets$ $attribution$
                \State Process auto-differentation on $value$
                \State \Return list of gradient values for each bit
            \EndProcedure
        \end{algorithmic}
        \end{footnotesize}
    \end{algorithm}
\end{figure}

For computing the attribution of each bit in the target tensor, we re-implemented the definition of floating point over 32 bits \cite{IEEEStandardFloatingPoint2019}, computing the real value out of the bit list. By carrying out the computations directly, we can exploit the auto-differentiation capabilities of different libraries, e.g., PyTorch, to compute the gradients based on the importance of the tensor, and use them as importance for each bit.
We use the gradient as attribution as it can be seen as the approximation of the linear coefficient for a weighted sum across all the bits generating the tensor value.
The algorithm is shown in Algorithm \ref{algorithm:methodology_bit_gradient}. There are differences with the implementation in the IEEE standard, however, they revolve around handling exceptions and special cases: we opted for not using directly the \titlecaseabbreviation{nan} and $\pm\infty$ values, as doing so would affect the computed gradients. Hence, we use flags to represent these special cases, as shown in Lines \ref{algorithm:line:flag_start}-\ref{algorithm:line:flag_end}, and then we multiply them by constants to allow for the auto-differentiation algorithm to compute the gradients. In the case of $\pm\infty$, we normalize by the value, and then multiply by the maximum non-infinite number divided by 33, so that even if the value is close to infinity it would not be exactly $\pm\infty$. This process is shown in Line \ref{algorithm:line:flag_multiply_inf}.

\subsection{Fault Model}

After computing the attributions, the fault model uses the attributions for the selected injection target, neuron weight or neuron output, as weights for the sampling algorithm.
Depending on the application and the user, the fault model can also incorporate a percentage of random uniform fault samples, to increase coverage while still providing higher precision with the importance sampling.
The fault model is easily customizable, so that extra information can be added in how the faults are sampled.

\subsection{Fault Injection Framework}

The faults sampled by the fault model are then injected through the use of a fault injection framework, which compares the baseline accuracy over the whole testing dataset of the model-under-test with the accuracy when using the faults.
The choice of the fault injection framework is dictated by the campaign requirements in terms of resources and customizability, but \emphasizedworkname{} is independent of the fault injection framework as long as it can receive the fault location as input.

\subsection{Output}

The results of the injection are summarized in a SQLite database or CSV file, depending on the fault injection framework, which can be accessed and used for further visualization, as it will be shown in Section \ref{section:evaluation}.

\subsection{Application Example: Fault-Aware Training}

\begin{figure}[h]
    \centering
    \includegraphics[width=\linewidth]{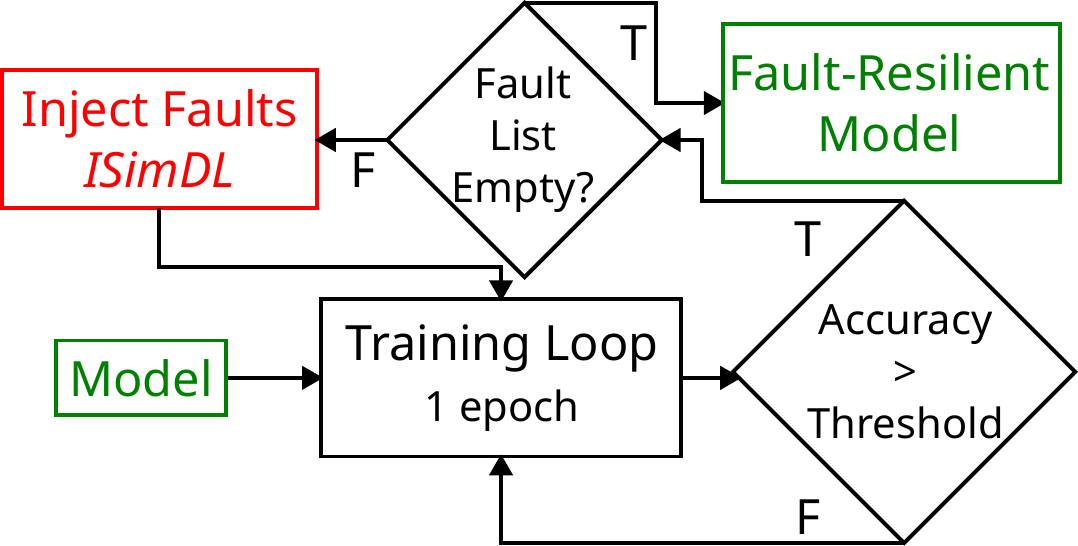}
    \caption{Methodology overview for \glsentrytitlecase{fat}{long}. The model, shown as input on the bottom left, is trained until a certain accuracy threshold is reached, and if so, faults are injected using \emphasizedworkname{}, while training continues until all faults are injected and the target accuracy is reached. The resulting model is fault-resilient against the injected faults, shown on the top right.}
    \label{figure:fat_methodology}
\end{figure}

\titlecaseabbreviation{fat} \cite{mahmoudPyTorchFIRuntimePerturbation2020} is a technique used to make neural network models more resilient to faults and the corresponding \titlecaseabbreviationpl{sdc} that might happen in real-world scenarios. Generally, faults are uniformly sampled from a database, or they follow some specific human-devised fault model, and they are integrated in the training loop, as shown in Figure \ref{figure:fat_methodology}. In our use-case, we use \emphasizedworkname{} to inject faults, showing it can be used to speed up the search for faults leading to \titlecaseabbreviationpl{sdc}.

\section{Experimental Setup}
\label{section:experimental_setup}

\begin{figure}[h]
    \centering
    \includegraphics[width=\linewidth]{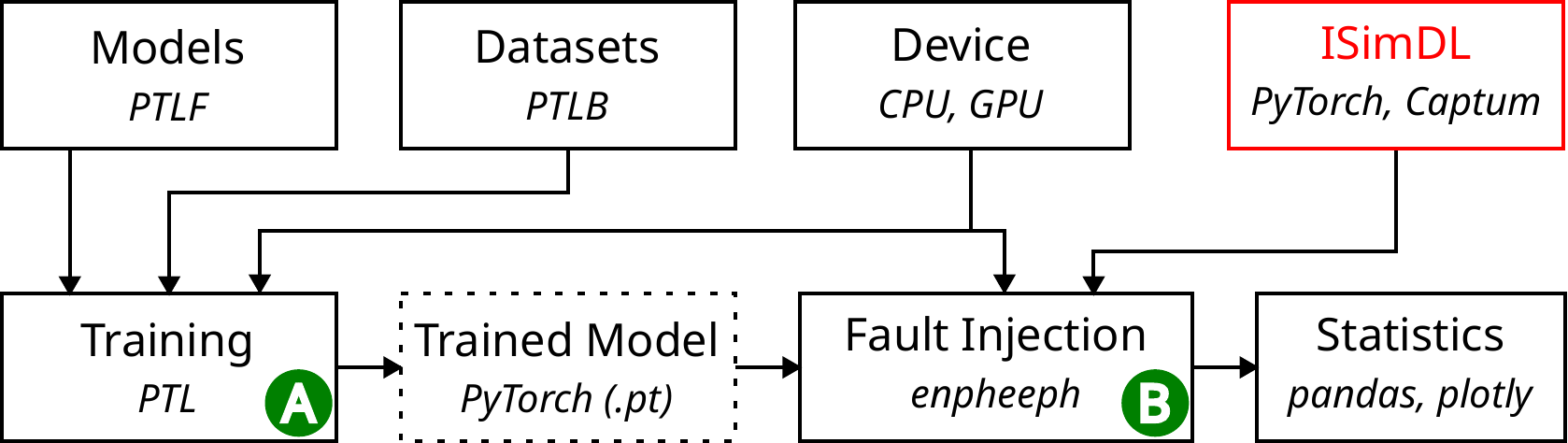}
    \caption{Experimental setup for evaluating \emphasizedworkname{}. Different models on different datasets are trained on GPU, and then injected using enpheeph as fault injection framework and \emphasizedworkname{} to generate the faults. The results are analyzed and plotted to be compared with random uniform fault sampling.}
    \label{figure:experimental_setup}
\end{figure}

To evaluate \emphasizedworkname{}, we use the experimental setup shown in Figure \ref{figure:experimental_setup}. First the models are trained on different datasets, as shown in \GoodDarkGreenCircled{A}, and they are then used for the fault injection campaign with \emphasizedworkname{} in \GoodDarkGreenCircled{B}. The technical details of the implementations are discussed in the following sub-sections.

\subsection{Hardware}

We employ 4 Nvidia RTX 2080 Ti GPUs \cite{GraphicsReinventedNVIDIA}, with AMD Ryzen Threadripper 2990WX CPU \cite{AMDRyzenThreadripper}, for training the selected models. The models are being read/written on SSD, to avoid the speed limit of hard drives.
We use Manjaro Linux as operating system.

\subsection{Software}

We train \gls{sota} \titlecaseabbreviation{dnn} models, VGG-11 \cite{simonyanVeryDeepConvolutional2015} and ResNet18 \cite{heDeepResidualLearning2016}, on common benchmark datasets, CIFAR10 \cite{Krizhevsky2009LearningML} and GTSRB \cite{stallkampManVsComputer2012}, using \gls{sota} hyperparameter settings and techniques, such as the Adam optimizer \cite{kingmaAdamMethodStochastic2015} with learning rate \num{0.001}, 30 epochs of training with batch size 64, and validation. These models are trained using PyTorch Lightning \cite{Falcon_PyTorch_Lightning_2019} and its auxiliary libraries, PyTorch Lightning Bolts \cite{falcon2020framework} and PyTorch Lightning Flash \cite{PyTorchLightningLightningflash2022}.
We implement \emphasizedworkname{} using PyTorch \cite{paszkePyTorchImperativeStyle2019} and the enpheeph fault injection framework \cite{colucciEnpheephFaultInjection2022a}, to inject the faults and compare them with random uniform sampling. Additionally, we use Captum  \cite{kokhlikyanCaptumUnifiedGeneric2020} for implementing the attribution algorithm for the neuron output.
The outcomes are recorded in CSV, analyzed with pandas \cite{the_pandas_development_team_2023_7549438} and plotted using Plotly \cite{hossainVisualizationBioinformaticsData2019}.

\subsection{Fault-Aware Training}

The setup for the \titlecaseabbreviation{fat} is similar to the general setup, however, the model is a custom smaller \titlecaseabbreviation{dnn} employing 2 convolutional layers with ReLU as activation, followed by 2 fully-connected layers with ReLU as activation. We employ a smaller model as it is easier to analyze the generated fault patterns compared to more complex models.
The model is trained on FashionMNIST \cite{xiaoFashionMNISTNovelImage2017} with the Adam optimizer, using the AMD CPU.

\subsection{Metrics}
\label{section:experimental_setup:subsection:metrics}

To measure the efficacy of \emphasizedworkname{} compared to random uniform sampling, we compute the rate of discovered \titlecaseabbreviationpl{sdc} over the total number of injected faults, computing the \ms{Precision} of \emphasizedworkname{}.
We use the statistical definition of precision:

\begin{equation}
    0 \leq \ms{Precision} = \frac{\ms{Positive_{True}}}{\ms{Positive_{True}} + \ms{Positive_{False}}} \leq 1
\end{equation}

where \ms{Positive} represents the elements which are recognized as part of the target set of elements by \emphasizedworkname{}, with \ms{True} meaning that they are part of the actual target set, and \ms{False} as they were wrongly categorized, and they are not part of the target set. The target set is defined based on the accuracy drop caused by the \titlecaseabbreviation{sdc} generated by the fault: when we write \glsxtrshort{sdc}\num{0.05} we mean all the faults that lead to a \titlecaseabbreviation{sdc} generating an accuracy drop of \num{5}\% compared to the fault-free baseline.

We also define the \ms{Recall}, as the number of faults leading to a predetermined \titlecaseabbreviation{sdc} threshold drop over the total number of possible faults in the whole search space:

\begin{equation}
    0 \leq \ms{Recall} = \frac{\ms{Positive_{True}}}{\ms{Positive_{True}} + \ms{Negative_{False}}} \leq 1
\end{equation}

where $\ms{Negative_{False}}$ represents all the possible non-\titlecaseabbreviation{sdc} faults, and therefore must be either estimated or enumerated over the whole space of possible faults. We use the \ms{Recall} metric only for comparison with \titlecaseabbreviation{sota}.

\subsection{Experiments}
\label{section:experimental_setup:subsection:experiments}

We run the whole set of experiments for each of the 4 model-dataset configurations which have been trained:
\begin{itemize}
    \item VGG11-CIFAR10;
    \item VGG11-GTSRB;
    \item ResNet18-CIFAR10;
    \item ResNet18-GTSRB.
\end{itemize}

Each configuration undergoes the same set of experiments, to understand how the importance sampling through the different attributions compares to the random uniform sampling. To this end, we additionally implement two extra bit attribution algorithms:

\begin{itemize}
    \item linear bit-weighting: where each bit is weighted linearly starting from 1, so the \titlecaseabbreviation{lsb} in position 0 has a weight of 1, and the \titlecaseabbreviation{msb} in position 31 has a weight of 32, as in $\ms{weight} = i + 1, \forall i = 0, \ldots, 31 $;
    \item exponential bit-weighting: each bit is weighted in the form $2^i$, where $i$ is the index of the bit, starting from 0 with the \titlecaseabbreviation{lsb} and reaching 31 for the \titlecaseabbreviation{msb}, as in $\ms{weight} = 2^i, \forall i = 0, \ldots, 31 $.
\end{itemize}

\begin{table}[h]
    \caption{The short codes used to describe the different experiments to evaluate \emphasizedworkname{}. Each code is a combination of its two-level column and row locations. \emph{Target} refers to the injection target, \emph{Neuron} to the algorithm used to sample the neuron to be injected, using importance or with random uniform, \emph{Bit} represents the attribution used for sampling the bit to be injected.}
    \label{table:experiment_legend}
    \centering
    \setlength{\tabcolsep}{2pt}
    \resizebox{\linewidth}{!}{%
        \begin{tabular}{cc|cc|cc|}
\cline{3-6}
                      & Target               & \multicolumn{2}{c|}{\textbf{Output}} & \multicolumn{2}{c|}{\textbf{Weight}} \\ \cline{3-6} 
 &
  Neuron &
  \multicolumn{1}{c|}{\textbf{Importance}} &
  \textbf{Random} &
  \multicolumn{1}{c|}{\textbf{Importance}} &
  \textbf{Random} \\ \cline{2-6} 
\multicolumn{1}{c|}{\multirow{4}{*}{Bit}} &
  \textbf{Gradient} &
  \multicolumn{1}{c|}{GBINo} &
  GBRNo &
  \multicolumn{1}{c|}{GBINw} &
  GBRNw \\ \cline{2-6} 
\multicolumn{1}{c|}{} & \textbf{Exponential} & \multicolumn{1}{c|}{EBINo}  & EBRNo  & \multicolumn{1}{c|}{EBINw}  & EBRNw  \\ \cline{2-6} 
\multicolumn{1}{c|}{} & \textbf{Linear}      & \multicolumn{1}{c|}{LBINo}  & LBRNo  & \multicolumn{1}{c|}{LBINw}  & LBRNw  \\ \cline{2-6} 
\multicolumn{1}{c|}{} & \textbf{Random}      & \multicolumn{1}{c|}{RBINo}  & RBRNo  & \multicolumn{1}{c|}{RBINw}  & RBRNw  \\ \cline{2-6} 
\end{tabular}
    }
\end{table}

These alternative bit attribution algorithms are different approximations of the bit gradient algorithm, depending on the amount of overhead that is tolerable in the fault injection campaign.
Therefore, we use the codes shown in Table \ref{table:experiment_legend} to represent each experiment in Section \ref{section:evaluation}. As an example, we can choose the codes \textbf{GBINo} and \textbf{RBRNw}, with the former meaning we have used the \textbf{G}radient \textbf{B}it attribution together with the \textbf{I}mportance attribution sampling for the \textbf{N}euron \textbf{o}utput, while the latter stands for \textbf{R}andom \textbf{B}it with \textbf{R}andom uniform sampling for the \textbf{N}euron \textbf{w}eight.

Each experiment is run for \num{5} times with the \num{5} different fixed seeds across all the Python libraries, to improve reproducibility of the results and reduce non-deterministic operations, while also averaging out any out-of-distribution irregularities, and each single experiment per seed lasts for exactly \num{24} hours, covering roughly \num{10000} faults. The achieved standard deviation is roughly 0.05 for the precision metric.

\subsubsection[Fault Aware Training]{\glsentrytitlecase{fat}{long}}
\label{section:experimental_setup:subsection:experiments:subsubsection:fault_aware_training}

Regarding the \titlecaseabbreviation{fat}, we employ a similar setup for the baseline training, training the network for 30 epochs with batch size 64, using the Adam optimizer with \num{0.01} learning rate.
After training the model for the baseline accuracy measurements, we retrain it from scratch, and after each epoch we run 3 fault injection simulations with a different random seed each, and for each simulation we measure the time required to reach different \titlecaseabbreviation{sdc} thresholds for 3 times in a row, to reduce the effect of initialization in the experiments. Each simulation covers all the experiments shown in Table \ref{table:experiment_legend}. The \titlecaseabbreviation{sdc} thresholds range from \num{0.00} to \num{0.90} in \num{0.05} steps, as the base accuracy of the trained model is \num{90}\%, therefore it would not be possible to reach below-zero accuracy.
Finally, we select the first 5 faults from the \emphasizedworkname{} simulations, for both gradient bit weighting with importance sampling for neurons and random uniform sampling, and we inject them after 5 epochs of training, to confirm possibility of increasing the \titlecaseabbreviation{dnn} robustness to the chosen faults by means of training techniques. The results will be shown in Section \ref{section:evaluation:subsection:fault_aware_training}.

\section{Evaluation}
\label{section:evaluation}

We evaluate \emphasizedworkname{} based on the metrics and experiments mentioned in Sections \ref{section:experimental_setup:subsection:metrics} and \ref{section:experimental_setup:subsection:experiments}.

\subsection[Comparison with state of the art]{Comparison with \glsentrylong{sota}}

\begin{table}[h]
    \caption{Comparison between \emphasizedworkname{} and other \gls{sota} fault models: \emphasizedworkname{} does not require a-priori knowledge, and achieves slightly lower precision but in much fewer number of samples, while still being capable of reaching \num{100}\% recall.}
    \label{table:evaluation_sota}
    \centering
    {\renewcommand{\arraystretch}{1.4}
    \setlength{\tabcolsep}{2pt}
    \resizebox{\linewidth}{!}{%
        \begin{tabular}{c|c|c|c|}
\cline{2-4}
\multicolumn{1}{l|}{}                             & \textbf{ISimDL} & \textbf{BinFI \cite{chenBinFIEfficientFault2019}} & \textbf{AVFI \cite{jhaMLBasedFaultInjection2019}}  \\ \hline
\multicolumn{1}{|c|}{\textbf{A-priori Knowledge}} & \greenxmark             & \redcheckmark         & \redcheckmark         \\ \hline
\multicolumn{1}{|c|}{\textbf{Precision (\%)}}     & \num{59.4}            & \num{78}             & \num{82}             \\ \hline
\multicolumn{1}{|c|}{\textbf{\# Samples for Precision}} & \textless{}\num{100} & $\frac{1}{5}$ Full Campaign & $\sim$\num{100000} \\ \hline
\multicolumn{1}{|c|}{\textbf{Recall (\%)}}        & \num{100}             & \num{99}             & \textless{}\num{100} \\ \hline
\end{tabular}%
    }
    }
\end{table}

We compare \emphasizedworkname{} with other \gls{sota} fault models, namely BinFI \cite{chenBinFIEfficientFault2019} and AVFI \cite{jhaMLBasedFaultInjection2019}. We get the other models' results from their own papers. The comparison is shown in Table \ref{table:evaluation_sota}.
BinFI has very high recall, however it requires running $\frac{1}{5}$ of the total possible faults occurring in the system-under-test. AVFI reaches a high precision of \num{82}\%, however it still needs to execute around \num{100000} faults. Both models require external a-priori knowledge, assuming the data type used by the model and its characteristics in the former, while assuming the system configuration and the importance of the different components in the latter.

On the other hand, \emphasizedworkname{} requires no a-priori knowledge for running the importance algorithms, and while its precision is slightly lower than \gls{sota} with \num{59.4}\%, it is the average of multiple experiments and not a single campaign as in the \gls{sota}, and it requires less than 100 samples to reach it, while at the same time having \num{100}\% recall if the fault injection campaign runs for a longer time.

\subsection{Attribution Efficacy}
\label{section:evaluation:subsection:attribution_efficacy}

First, we focus our attention on the attribution algorithms which have been developed as part of \emphasizedworkname{}. We compare the precision of the fault injection campaigns when running all the possible experiments as shown in Table \ref{table:experiment_legend}.
Therefore, we can understand whether the methodology applied by \emphasizedworkname{} is better than random uniform sampling, while still not relying on external knowledge when selecting the faults.

\subsubsection{Neuron Output}

\begin{figure}[h]
    \centering
    \includegraphics[width=\linewidth]{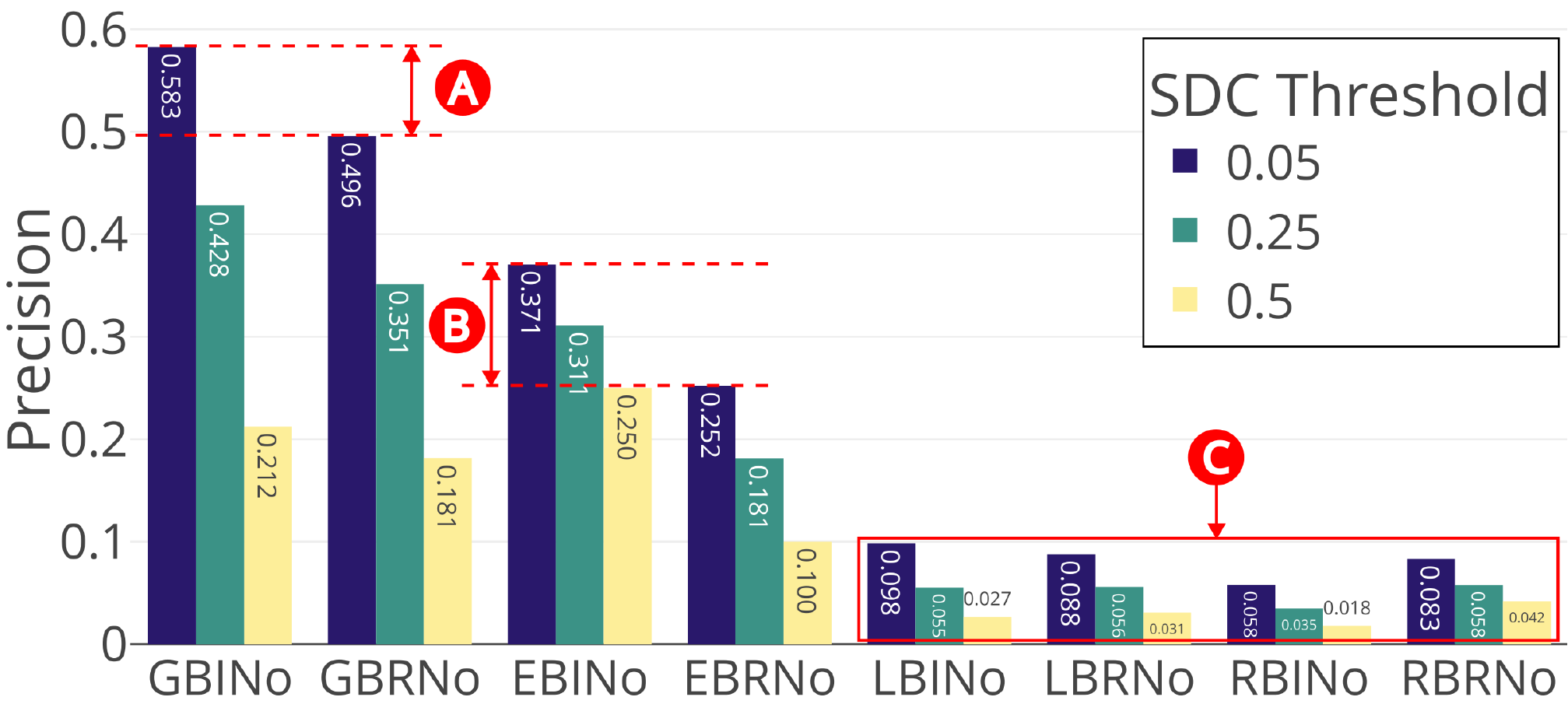}
    \caption{Average precision of \emphasizedworkname{} across different \titlecaseabbreviation{sdc} thresholds when running different neuron output experiment configurations on ResNet18-GTSRB. Importance sampling for neurons increases the precision for bit-weighting algorithms which are closer to the real importance of the bits, as in gradient and exponential weights, while having less impact with random uniform and linear weights.}
    \label{figure:histogram_activation}
\end{figure}

We start with the analysis of the neuron output simulations on ResNet18-GTSRB. The results for the average precision are shown in Figure \ref{figure:histogram_activation}. We can see importance sampling with neuron output attribution achieving higher precision than with random uniform sampling of neurons, hence highlighting the importance used by \emphasizedworkname{} as a working algorithm. As pointed by \GoodRedCircled{A}, there is a \num{0.087} difference in precision between \textbf{GBINo} and \textbf{GBRNo} at \titlecaseabbreviation{sdc}\num{0.05}, which further increases to \num{0.121} when considering \textbf{EBINo} and \textbf{EBRNo}, as shown by \GoodRedCircled{B}. This is due to the capability of selecting the neurons which have the most effect on the \titlecaseabbreviation{dnn} execution, hence leading to wider accuracy drops when generating \titlecaseabbreviationpl{sdc}. However, importance sampling for neurons does not perform as well if the target bit is randomly or linearly chosen, as shown by \GoodRedCircled{C}: this can be attributed to the effects of using random uniform sampling and linear weighting for the target bit, which makes the effects of the fault too much dependent on the chosen bit rather than the chosen neuron output, hence limiting the efficacy of the importance attribution sampling. \emph{This highlights the requirement of employing attributions both at the tensor and at the bit level to achieve optimal results, as developed in \workname{}}.

\begin{figure}[h]
    \centering
    \includegraphics[width=\linewidth]{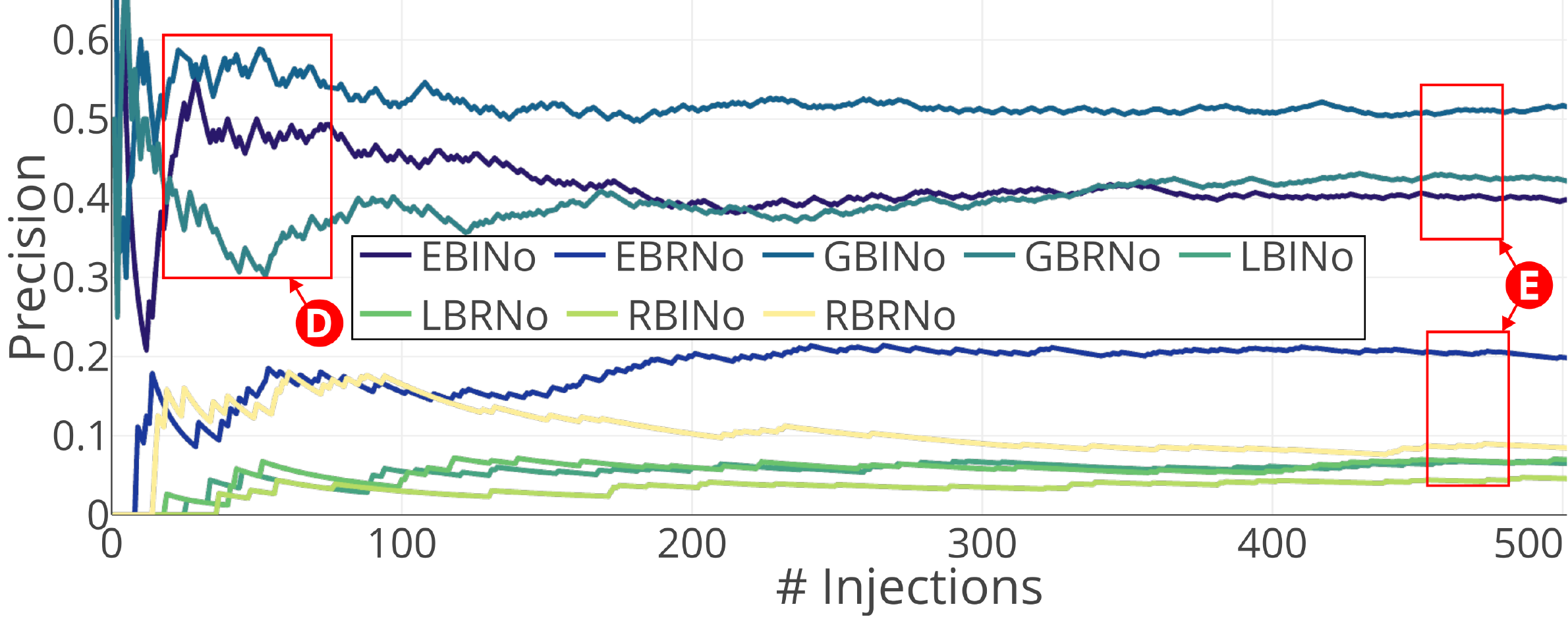}
    \caption{Average precision of \emphasizedworkname{} over number of samples for different neuron output experiment configurations with ResNet18-GTSRB. We can notice traces employing importance sampling for neurons having higher precision in the first 100 samples, decreasing to the steady-state values afterwards.}
    \label{figure:line_activation}
\end{figure}

We then analyze the same experiments on ResNet18-GTSRB but along the number of samples, shown in Figure \ref{figure:line_activation}. Each line represents the precision with increasing number of samples, and there are some interesting patterns: with a low number of samples, importance neuron sampling has higher precision than the steady-state average found in the previous Figure \ref{figure:histogram_activation}, as pointed by \GoodRedCircled{D} for the \textbf{GBINo} and \textbf{EBINo} traces. It then decreases to the steady-state value, while for random neuron sampling in \textbf{GBRNo} the behavior is opposite, being lower and then increasing to the steady-state value. These steady-state values are reached fairly soon in the campaign, as shown by \GoodRedCircled{E} at $\sim$\num{450} samples. These two groups, one at the top and one at the bottom, highlight yet again the choice of the bit weighting algorithm to be paramount for achieving high precision, together with the use of importance sampling for neurons, as linear and random uniform bit weights have precision lower than \num{0.10}.

\subsubsection{Neuron Weight}

\begin{figure}[h]
    \centering
    \includegraphics[width=\linewidth]{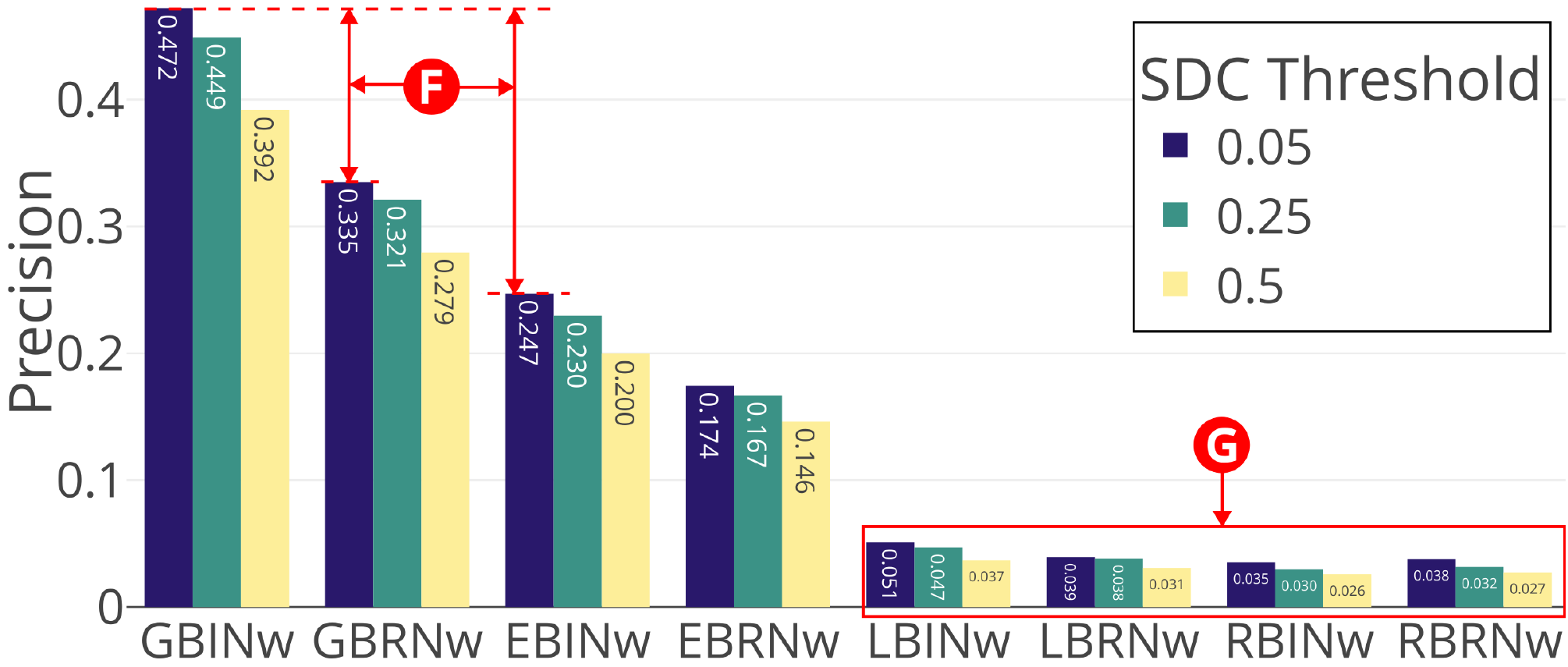}
    \caption{Average precision of \emphasizedworkname{} across different neuron weight experiment configurations on ResNet18-CIFAR10. The drop across experiment configurations is higher when employing a different bit weighting technique, rather than switching from importance sampling to random uniform sampling for neurons.}
    \label{figure:histogram_weight}
\end{figure}

We repeat a similar in-depth analysis for neuron weight injections on ResNet18-CIFAR10 using \emphasizedworkname{}. As we can see in Figure \ref{figure:histogram_weight}, both importance sampling for neuron and bit attribution are important when running fault injection campaigns. As pointed by \GoodRedCircled{F}, there is a bigger drop when going from \textbf{GBINw} to \textbf{EBINw} than going from \textbf{GBINw} to \textbf{GBRNw}. This highlights the importance of choosing the appropriate bit weighting technique, and with support from importance sampling for neurons the results can be further enhanced. This is further proved by noticing how linear bit weights and random uniform sampling for bits have much lower precision, even when using importance sampling for neurons, as pointed by \GoodRedCircled{G}. Therefore, the choice of the bit weighting technique is important in achieving high precision early on in the fault injection campaign.

\begin{figure}[h]
    \centering
    \includegraphics[width=\linewidth]{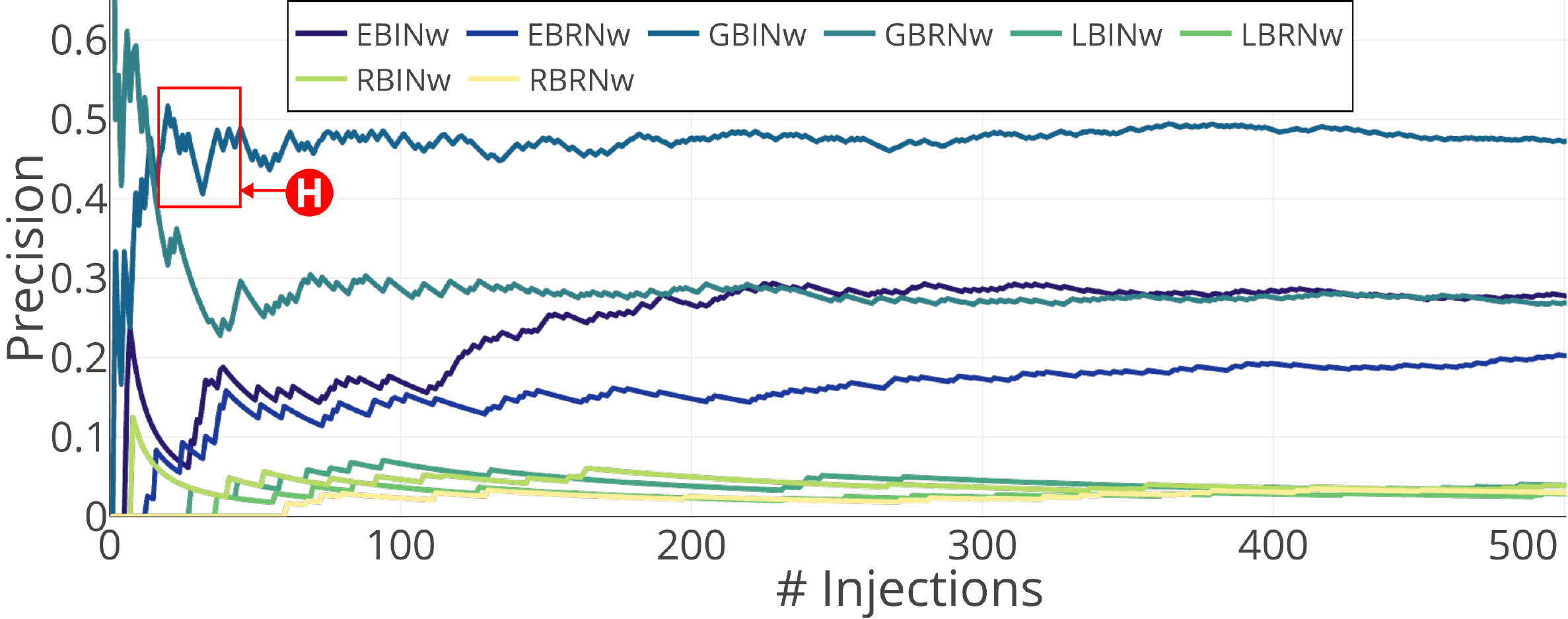}
    \caption{Average precision of \emphasizedworkname{} over number of samples for different neuron weight experiment configurations on ResNet18-CIFAR10. Importance sampling for neurons provide a precision value close to the steady-state value as soon as 50 samples after the start of the fault injection campaign, showing \emphasizedworkname{} can achieve high precision early in the campaign.}
    \label{figure:line_weight}
\end{figure}

Additionally, we analyze the evolution of precision of \emphasizedworkname{} running on ResNet18-CIFAR10 over an increasing number of samples, plotted in Figure \ref{figure:line_weight}. This plot shows steady-state results similar to the histograms in Figure \ref{figure:histogram_weight}. However, as pointed by \GoodRedCircled{H}, we can notice how the average precision for \textbf{GBINw} reaches the steady-state value in less than 50 samples. This reinforces the value of importance sampling with \emphasizedworkname{} for achieving high precision early in the fault injection campaign. 

\subsection{Comparison across different configurations}
\label{section:evaluation:subsection:different_configurations}

\begin{figure*}[h]
    \centering
    \includegraphics[width=\textwidth]{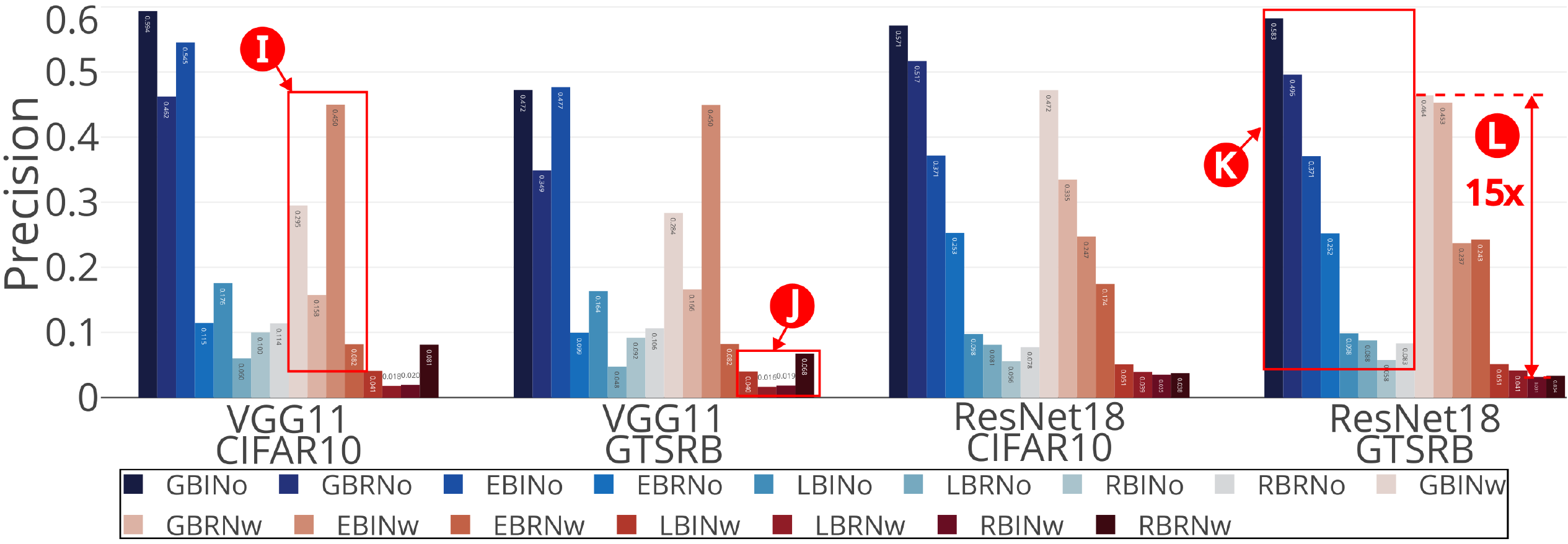}
    \caption{Average precision of \emphasizedworkname{} across different neuron weight experiment configurations on all the model-dataset pairs analyzed in this work. The average precision achieved by \emphasizedworkname{} is up to \num{15}$\times$ higher than the precision achieved by using random uniform sampling for both neuron and bit.}
    \label{figure:alltogehter_histogram}
\end{figure*}

Here, in Figure \ref{figure:alltogehter_histogram}, we show the average precision for all the different experiment configurations across all the possible model-dataset pairs. \emphasizedworkname{} has lower efficacy on the VGG11 model, as exponential bit weights achieve higher precision than gradient-based ones, as pointed by \GoodRedCircled{I}: this is related to the different internal architecture of the VGG model series compared to ResNet, and is dependent on the exact implementation of the importance sampling algorithm.
This effect is further extended as highlight by pointer \GoodRedCircled{J}, where the configuration \textbf{RBRNw} has higher precision than all the other ones up to \textbf{LBINw}.
For ResNet18-GTSRB, we have the results pointed by \GoodRedCircled{K} analyzed in deeper details in Figure \ref{figure:histogram_activation}. We see that the choice of importance sampling for neuron and the bit weighting technique increases the average precision during the fault injection campaign. In the case of weight injection, \emphasizedworkname{} with the \textbf{GBINw} experiment configuration reaches up to \num{15}$\times$ higher precision than the \textbf{RBRNw} experiment configuration, as shown by \GoodRedCircled{L}.  

\subsection[Fault Aware Training]{\glsentrytitlecase{fat}{long}}
\label{section:evaluation:subsection:fault_aware_training}

\begin{figure}[h]
    \centering
    \includegraphics[width=\linewidth]{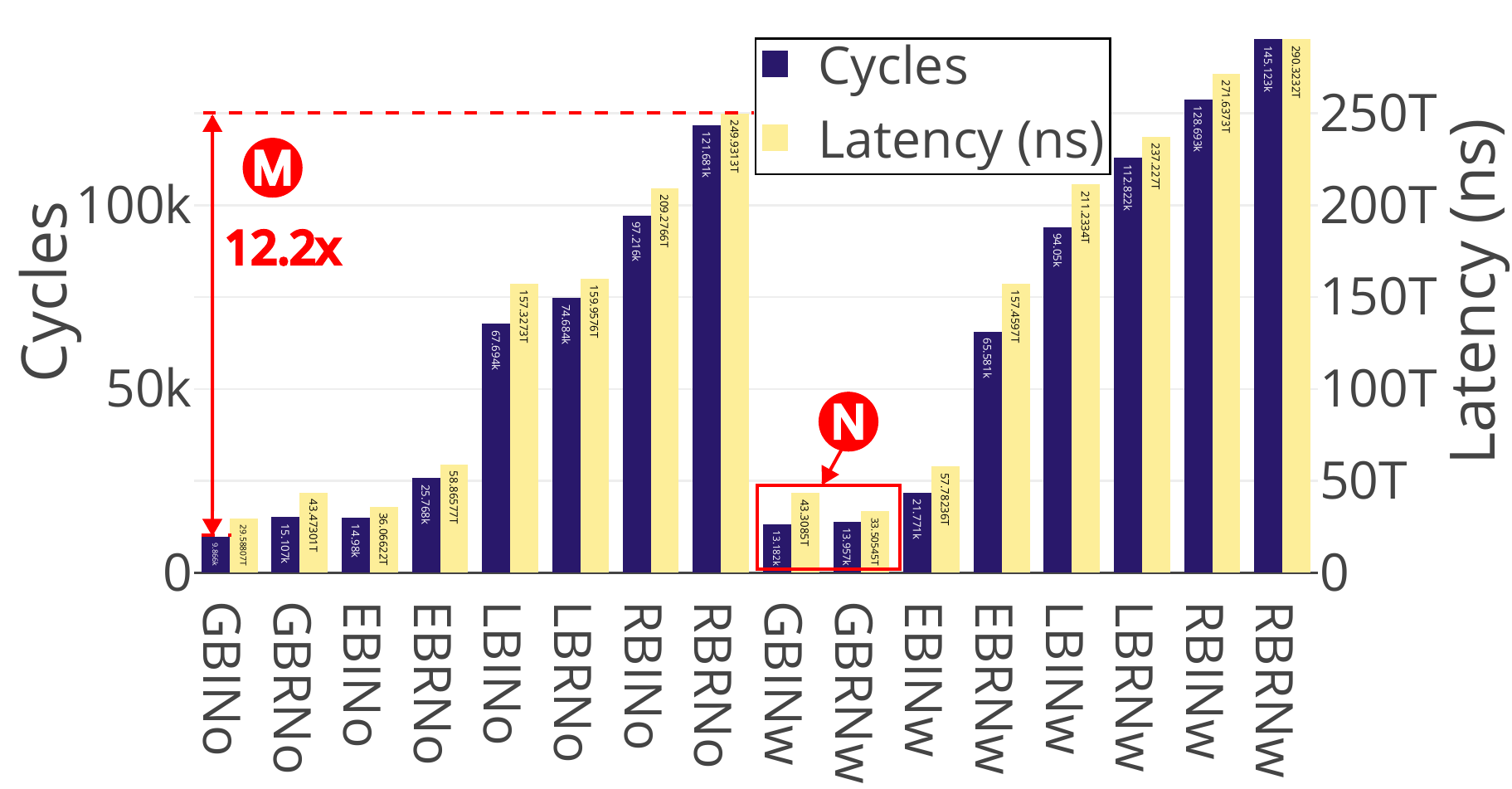}
    \caption{Comparison of average latency across different experiment configurations for achieving \num{3} consecutive critical faults with different \titlecaseabbreviation{sdc} thresholds. We can notice how gradient bit weighting outperforms other methods, and importance sampling for neurons further reduces the required number of cycles, reaching to \num{12.2}$\times$ lower latency.}
    \label{figure:fat_latency}
\end{figure}

We now analyze the usage of \emphasizedworkname{} in a real-world use case scenario, \glsentrytitlecase{fat}{long}. We analyze the time and cycle latency required to gather 3 faults leading to different \titlecaseabbreviation{sdc} thresholds, as defined in Section \ref{section:experimental_setup:subsection:experiments:subsubsection:fault_aware_training}. We can notice from the histogram of the latency in nanoseconds and the number of cycles in Figure \ref{figure:fat_latency}, that \textbf{GBINo} has \num{12.2}$\times$ lower number of cycles than \textbf{RBRNo}, pointed by \GoodRedCircled{M}, therefore showcasing that \emphasizedworkname{} can provide fault injection speed ups compared to random uniform sampling.
Additionally, in some cases, the computation of the attribution can increase the overhead of the fault injection, as pointed by \GoodRedCircled{N}, showing that \textbf{GBINw} has more latency in nanoseconds than latency in number of cycles than \textbf{GBRNw}.

\begin{table}[h]
    \caption{Accuracy for the training of the custom \titlecaseabbreviation{dnn} on FashionMNIST using \titlecaseabbreviation{fat}. The final results show that training while injecting the selected faults provide better coverage of possible \titlecaseabbreviationpl{sdc}, while reaching a similar baseline training accuracy. However, when using the faults generated with \textbf{RBRNo}, the model is similarly affected by the critical faults found using \textbf{GBINo}, making \emphasizedworkname{} more effective in selecting proper critical faults to train against with \titlecaseabbreviation{fat}.}
    \label{table:fat_training}
    \centering
    \setlength{\tabcolsep}{2pt}
    \begin{tabular}{c|c|c|c|}
\cline{2-4}
 & \textbf{Fault-Free} & \textbf{\begin{tabular}[c]{@{}c@{}}GBINo\\ Faults\end{tabular}} & \textbf{\begin{tabular}[c]{@{}c@{}}RBRNo\\ Faults\end{tabular}} \\ \hline
\multicolumn{1}{|c|}{\textbf{\begin{tabular}[c]{@{}c@{}}Baseline\\ Accuracy (\%)\end{tabular}}}  & 90.625  & 83.240 & 89.240 \\ \hline
\multicolumn{1}{|c|}{\textbf{\begin{tabular}[c]{@{}c@{}}GBINo FAT\\ Accuracy (\%)\end{tabular}}} & 90.625  & \red{90.063} & 90.063 \\ \hline
\multicolumn{1}{|c|}{\textbf{\begin{tabular}[c]{@{}c@{}}RBRNo FAT\\ Accuracy (\%)\end{tabular}}} & 90.6250 & \red{83.063} & 89.625 \\ \hline
\end{tabular}
\end{table}

After selecting the first 5 faults from the sampled ones in the previous experiment for both the \textbf{GBINo} and \textbf{RBRNo} experiment configurations, we try training the test network with them, and we report the different accuracies in Table \ref{table:fat_training}: the final post-\titlecaseabbreviation{fat} accuracy is very close to the original one for both \textbf{GBINo} and \textbf{RBRNo}, as well as the accuracy when injected with the set of faults they were trained against. However, as the faults genered by \textbf{GBINo} were critical, this leads to an overall more resilient model, while for \textbf{RBRNo} this does not hold. Therefore, when the \textbf{RBRNo}-trained model is injected with the \textbf{GBINo}-generated faults, the accuracy is as low as for the baseline model, with a drop from the baseline more than \num{4}$\times$ larger than for \textbf{GBINo}, showing that \emphasizedworkname{} can generate critical faults useful in improving the \titlecaseabbreviation{dnn} model resilience in practical applications like \titlecaseabbreviation{fat}.

\section{Conclusion}
\label{section:conclusion}
\glslocalresetall
In the nano-era, devices have become more susceptible to transient faults. As \titlecaseabbreviation{dl} systems became common in many applications, it is of paramount importance to guarantee their safety against all possible faults. Current \gls{sota} techniques for fault injection testing are inefficient and time-expensive.
We develop the \emphasizedworkname{} methodology, developing and implementing multiple importance sampling algorithms, increasing the weights of the most important neurons and bits when sampling for specific fault locations.
We employ \emphasizedworkname{} to compute critical faults of different \titlecaseabbreviation{dnn} model-dataset pairs, namely VGG11 and ResNet18 for the models and CIFAR10 and GTSRB for the datasets. In this analysis, we show that importance sampling for neurons and bits leads to up to \num{15}$\times$ higher precision in selecting critical faults against random uniform sampling, achieving this precision in less than \num{100} faults. This is in contrast with \gls{sota} fault models, which require external a-priori knowledge and hundreds of thousands of faults to achieve high precision.
We additionally test \emphasizedworkname{} in a real-world use case scenario, \titlecaseabbreviation{fat}, to generate faults and use them while training a \titlecaseabbreviation{dnn} model to make it resilient to the generated faults. When compared to random uniform sampling, \emphasizedworkname{} generates more critical faults with up to \num{12}$\times$ lower overhead.
We have shown \emphasizedworkname{} to be more effective than random uniform sampling and various \gls{sota} fault models, future work directions will focus on better analyzing the resilience of different models and implementations to develop optimal fault mitigation techniques.


\printbibliography


\end{document}